\documentclass[letterpaper]{article} 
\usepackage[submission]{aaai23}  
\usepackage{times}  
\usepackage{helvet}  
\usepackage{courier}  
\usepackage[hyphens]{url}  
\usepackage{graphicx} 
\usepackage{natbib}  
\usepackage{caption} 
\usepackage{algorithm}
\usepackage{algorithmic}
\usepackage{kotex}
\usepackage{multirow}
\usepackage{makecell}
\usepackage{amsmath}
\usepackage{bbding}
\usepackage{pifont}
\newcommand{\cmark}{\ding{51}}
\newcommand{\xmark}{\ding{55}}

\usepackage{newfloat}
\usepackage{listings}
\DeclareCaptionStyle{ruled}{labelfont=normalfont,labelsep=colon,strut=off} 
\floatstyle{ruled}
\newfloat{listing}{tb}{lst}{}
\floatname{listing}{Listing}
\title{LEAT: Towards Robust Deepfake Disruption in Real-World Scenarios via \\ Latent Ensemble Attack}
\author{
    Joonkyo Shim, Hyunsoo Yoon
}
\affiliations{
    Department of Industrial Engineering, Yonsei University, Seoul, Republic of Korea\\
    \{shimjk, hs.yoon\}@yonsei.ac.kr
}

\begin{document}

\maketitle

\begin{abstract}
Deepfakes, malicious visual contents created by generative models, pose an increasingly harmful threat to society. To proactively mitigate deepfake damages, recent studies have employed adversarial perturbation to disrupt deepfake model outputs. However, previous approaches primarily focus on generating distorted outputs based on only predetermined target attributes, leading to a lack of robustness in real-world scenarios where target attributes are unknown. Additionally, the transferability of perturbations between two prominent generative models, Generative Adversarial Networks (GANs) and Diffusion Models, remains unexplored. In this paper, we emphasize the importance of \emph{target attribute-transferability} and \emph{model-transferability} for achieving robust deepfake disruption. To address this challenge, we propose a simple yet effective disruption method called \emph{Latent Ensemble ATtack (LEAT)}, which attacks the independent latent encoding process. By disrupting the latent encoding process, it generates distorted output images in subsequent generation processes, regardless of the given target attributes. This target attribute-agnostic attack ensures robust disruption even when the target attributes are unknown. Additionally, we introduce a \emph{Normalized Gradient Ensemble} strategy that effectively aggregates gradients for iterative gradient attacks, enabling simultaneous attacks on various types of deepfake models, involving both GAN-based and Diffusion-based models. Moreover, we demonstrate the insufficiency of evaluating disruption quality solely based on pixel-level differences. As a result, we propose an alternative protocol for comprehensively evaluating the success of defense. Extensive experiments confirm the efficacy of our method in disrupting deepfakes in real-world scenarios, reporting a higher defense success rate compared to previous methods.

\end{abstract}

\section{Introduction}

\begin{figure}[t]
\centering
\includegraphics[width=0.8\columnwidth]{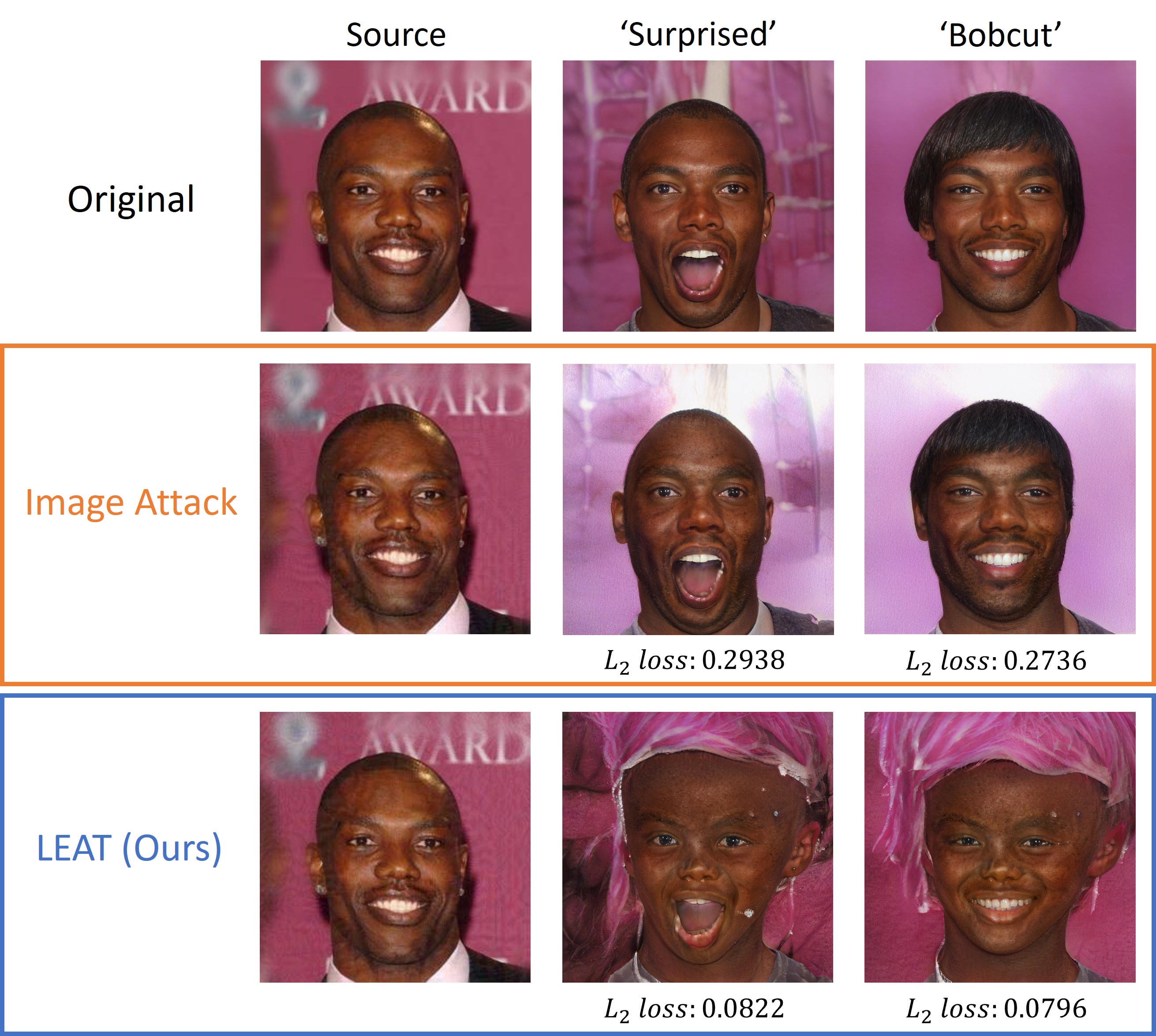}
\caption{Disrupted outputs of StyleCLIP in a gray-box scenario with unseen target attributes. While Image Attack shows higher $L_2$ loss between the generated images, it fails to disrupt the output. In contrast, our LEAT successfully disrupts the output in both cases.}
\label{fig1}
\end{figure}

With the remarkable success of generative models, such as Generative Adversarial Networks (GANs) and Diffusion Models, it has become increasingly feasible for anyone to create various realistic images and videos. Unfortunately, this advancement has given rise to a concerning issue in our society known as deepfakes, which involve the malicious use of fabricated visual content. Until now, the primary approach to combating deepfakes has been through passive defense mechanisms, which employ deepfake detection systems that assess the authenticity or manipulation of an image or video \cite{ijcai2020p476,tariq2021one,zhao2021multi}. However, these systems cannot completely prevent deepfakes since they only work after fake images have already spread over social media. To address this issue, an alternative approach called active defense has been proposed \cite{ruiz2020disrupting}. This method involves introducing human-imperceptible perturbations into deepfake models through adversarial attacks, resulting in the generation of distorted images. By incorporating perturbations in advance, it becomes possible to prevent an image from being used as a source for creating deepfakes.

To achieve effective disruption of deepfakes in real-world scenarios, two types of transferability are crucial: (1) target attribute-transferability and (2) model-transferability. Both of them are necessary to ensure that a single perturbation can cover all target attributes within a specific model and simultaneously affect different models. Previous methods \cite{ruiz2020disrupting,yeh2020disrupting} employ ensemble strategies, known as Image Attack, to achieve these transferabilities by averaging losses between generated images and considering all possible target attributes for each model. However, the emergence of text-driven manipulation models like StyleCLIP \cite{patashnik2021styleclip} presents a challenge in achieving target attribute-transferability due to unlimited manipulation possibilities. Consequently, handling all potential targets become infeasible. While generating perturbations assuming several known target attributes is possible, it does not guarantee their effectiveness in a gray-box scenario where the target attributes are unknown. In Figure \ref{fig1} (Image Attack), an ensemble strategy is used to attack five known target attributes, but it fails to effectively disrupt the output in gray-box scenarios, resulting in minor changes to the background. Additionally, the development of Diffusion Models presents another challenge in terms of model-transferability. Previous studies have targeted either GAN-based models or Diffusion-based models, but the possibility of simultaneously targeting both models remains unexplored.

In this work, our objective is to effectively achieve two aforementioned transferabilities. Firstly, we propose a Latent Ensemble Attack (LEAT) to achieve robust target attribute-transferability. Given that malicious individuals manually determine the selection of target attributes, it is critical to ensure robust disruption in both white-box scenarios, where the deepfake model and the target attributes are explicitly known, and gray-box scenarios, where target attributes are unknown. To achieve this, we divide deepfake models into two distinct processes: the latent encoding process, which encodes the semantic information of the input, and the generation process, which decodes the latent and target attributes to produce the desired output. Since the target attributes are only utilized in the generation process, we exploit the independence of the latent encoding process. Motivated by this, LEAT exclusively attacks the intermediate latent space, leaving the generation process unused. By disrupting the latent encoding process, subsequent disruption occurs in the generation process, irrespective of any specific target attribute. Consequently, LEAT achieves robust disruption in a gray-box scenarios, even against deepfake models that present unlimited target guidance. Unlike previous methods focusing on averaging the losses from every possible pair of output images, LEAT effectively achieves target attribute-transferability in a target attribute-agnostic manner, without generating any output image. Furthermore, LEAT dramatically reduces perturbation generation time compared to the Image Attack approach by forwarding only the latent encoding process and attacking the latent space of each model once.

Secondly, to achieve effective model-transferability, we propose a Normalized Gradient Ensemble, which is designed to obtain improved gradient directions for generating perturbations in the adversarial attack. Our ensemble strategy aggregates the gradient of each model while considering the scale difference among them. This ensures that all models significantly contribute during the ensemble process, enabling effective attacks on multiple models simultaneously. Unlike previous approaches that mainly focus on GAN-based face attribute manipulation models, our method targets all three categories of deepfakes, including face attribute manipulation, face swapping, and face reenactment. This encompasses both GAN-based and Diffusion-based models, which have substantially different structures. Additionally, we demonstrate that the perturbation generated by our method can be effectively applied to a black-box scenario, where we may even have no prior knowledge about the specific deepfake models involved.

Our contributions are summarized as follows:
\begin{itemize}
    \item We propose the Latent Ensemble Attack (LEAT), a method for achieving fully target attribute-agnostic deepfake disruption. LEAT focuses on attacking the latent encoding process without relying on specific target attributes, thus ensuring robust target attribute-transferability even in the gray-box scenarios, where the target attributes are unknown.
    \item We introduce the Normalized Gradient Ensemble, an ensemble strategy designed to achieve effective model-transferability by aggregating the gradients of target models. Our strategy demonstrates high scalability to deepfake models and encompasses all three categories of deepfake. This is the first approach that simultaneously targets both GAN-based and Diffusion-based models.
    \item Through comprehensive experiments, we demonstrate the effectiveness of our method in disrupting both the intermediate latent space and the output image in real-world scenarios. This includes white-box, gray-box, and even black-box scenarios, where the specific deepfake models are unidentified.
\end{itemize}

\begin{figure*}[tb!]
\centering
\includegraphics[width=1.0\textwidth]{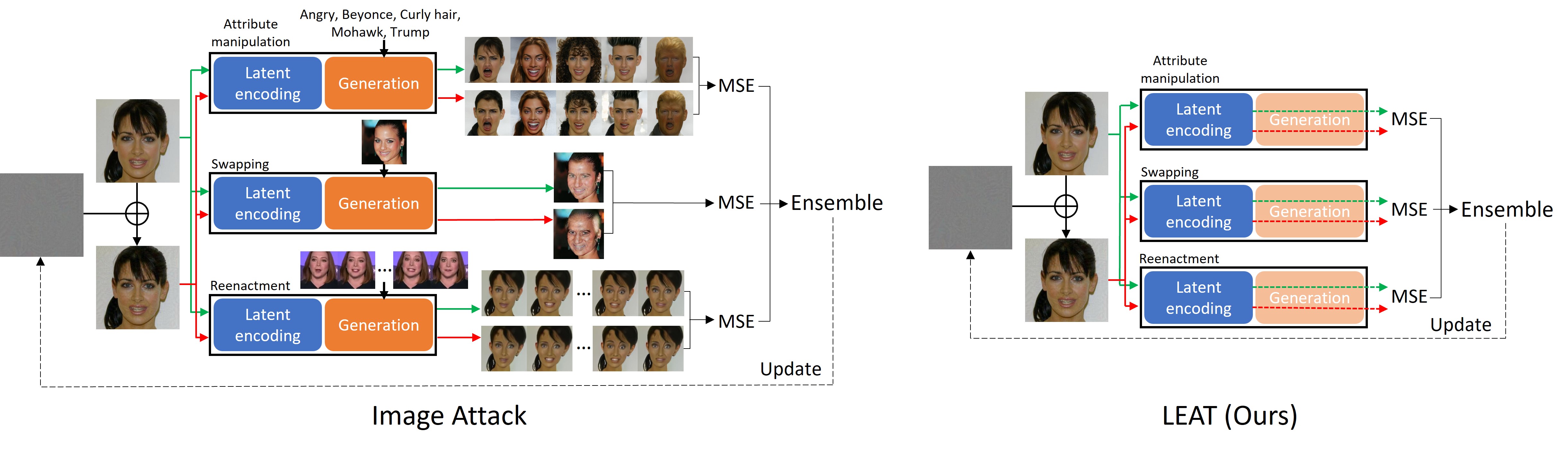}
\caption{The overall pipeline of Image Attack and our LEAT. The green arrows depict the original source and outputs, while the red arrows represent the perturbed source and disrupted outputs. In Image Attack, target attributes are employed in the generation process to generate desired output images. In contrast, LEAT does not utilize any of the target attributes, leaving the generation process unused.}
\label{fig_model}
\end{figure*}

\section{Related Works}

\subsection{Deepfake Methods}
Generative Adversarial Networks (GANs) have gained popularity for their ability to generate highly realistic images and videos. More recently, Diffusion Models have emerged as another prominent approach for generating visually appealing contents. However, any output produced by these models can be considered deepfakes when exploited maliciously. Deepfakes are typically classified into three main categories: face attribute manipulation, face swapping, and face reenactment. Face attribute manipulation \cite{choi2018stargan,he2019attgan,karras2019style,patashnik2021styleclip,preechakul2022diffusion} involves modifying specific facial attributes to achieve desired characteristics, such as altering hairstyles or expressions. Most notably, StyleCLIP \cite{patashnik2021styleclip} combines StyleGAN \cite{karras2019style,karras2020analyzing} and CLIP \cite{radford2021learning} to enable text-driven manipulation, expanding the range of potential manipulations beyond predetermined guidance. Face swapping \cite{li2019faceshifter,chen2020simswap}, on the other hand, involves extracting the face from a source image and seamlessly injecting it into the facial part of a target individual. Face reenactment \cite{pumarola2018ganimation,tripathy2020icface} focuses on transforming the source face to mimic the emotion and movements observed in the driving image or video. In our work, we simultaneously target four different models, covering all three categories mentioned above. By attacking these models collectively, we aim to address the challenges posed by a variety of deepfake generation techniques.

\subsection{Adversarial Attack}
Since the publication of \cite{42503}, which highlights the vulnerability of deep neural networks to imperceptible perturbations, various methods have been developed to generate adversarial examples specifically targeting classification models. \cite{43405} introduces a fast gradient sign method (FGSM) as one-step gradient attack to update each input pixel. \cite{45818} proposes the Iterative-FGSM, which performs gradient attack iteratively. \cite{madry2018towards} presents the projected gradient descent (PGD) methods, starting from randomly perturbed input to conduct a similar iterative attack. Furthermore, \cite{kos2018adversarial} applies adversarial attack to generative models and explores the possibility of attacking the latent vector of VAE-GAN, revealing the potential for latent attacks. Recently, \cite{liang2023adversarial} demonstrates the creation of adversarial examples for Diffusion Models.

\subsection{Deepfake Disruption}
Previous studies have introduced active defense techniques against deepfake models by employing adversarial perturbations \cite{ruiz2020disrupting,yeh2020disrupting}. \cite{guan2022defending} extends this approach to disrupt three different categories of deepfake models at the same time. \cite{huang2022cmua} proposes a universal adversarial watermark to enable cross-model and cross-image attacks. \cite{huang2021initiative,wang2022deepfake,aneja2022tafim} utilize neural networks to generate image-specific perturbations. These methods primarily focus on Image Attack, aiming to maximize or minimize the distance between the distorted output image and a specific target. There have also been attempts to disrupt the feature extraction module. \cite{tang2023feature} proposes a two-stage approach that first attacks the feature extractor of deepfake models and then performs end-to-end Image Attack. While they explore the impact of attacking the feature extraction module, they still rely on Image Attack to achieve better performance. Moreover, their approach assumes that the latent representation of each model is a feature map, which restricts their target models. In contrast, our method is the first approach that exclusively focuses on attacking the latent encoding process of each model, enabling a fully target attribute-agnostic attack. Moreover, our method is capable of disrupting multiple categories of deepfake models, regardless of the structural differences.

\section{Methods}
In this section, we introduce the mechanism of deepfake disruption. We then differentiate between the previous Image Attack approach and our proposed LEAT. Finally, we describe our Normalized Gradient Ensemble strategy.

\subsection{Disruption of Deepfake Models}
In general, the process of deepfake models can be formulated by:
\begin{equation*}
y=G(X,c),
\end{equation*}
where $X$ represents the source image, $c$ denotes the target attribute, $G$ is the generative model. To disrupt an image based on a specific model and target attribute, human-imperceptible perturbation $\eta$ is added to the source image to maximize the difference between the original output and the perturbed output by:
\begin{equation}\label{eq1}
\max_\eta L(G(X,c),G(X+\eta,c)),\,s.t.\,\|\eta\|_\infty\le\epsilon,
\end{equation}
where $\epsilon$ controls the magnitude of the perturbation. To obtain optimal $\eta$, FGSM \cite{43405} can be adopted as follows:
\begin{equation}\label{eq2}
\eta=\epsilon\,sign[\nabla_X L(G(X,c),G(X+\eta,c))].
\end{equation}
With I-FGSM \cite{45818} and PGD \cite{madry2018towards}, more powerful adversarial perturbation can be obtained through iterative gradient updates:
\begin{equation}\label{eq3}
X_{t+1}=clip(X_t+a\,sign[\nabla_{X_t} L(G(X_t,c),G(X_t+\eta,c))]),
\end{equation}
where $a$ is the step size for each iteration and $clip$ function keeps $X_t$ in the range [$X-\epsilon$,\,$X+\epsilon$] at every $t$. The key difference is that the source $X$ starts with random perturbation in PGD.

To disrupt multiple models and target attributes simultaneously, \cite{ruiz2020disrupting} and \cite{yeh2020disrupting} propose ensemble strategies to maximize the average distance between generated outputs across all possible models and target attributes. We consider their method Image Attack, formulated by: 
\begin{equation}\label{eq4}
\max_\eta\sum_{k}^{}\sum_{c_k}^{}L(G_k(X,c_k),G_k(X+\eta ,c_k)),\,s.t.\,\|\eta\|_\infty\le\epsilon,
\end{equation}
where $G_k$ and $c_k$ denote the generative models and their corresponding target attributes, respectively. As illustrated in Figure \ref{fig_model}, the Image Attack first generates original and perturbed outputs for all known target attributes of each model and then aggregates the mean squared error (MSE) loss between them. Note that each deepfake model can have different types and numbers of target attributes.

\begin{algorithm}[t]
\caption{LEAT with Normalized Gradient Ensemble}
\label{alg:algorithm}
\textbf{Input}: $X$ (facial image), $E_1,...,E_K$ (latent encoding modules), $T$ (number of iterations), $a$ (step size), $\epsilon$ (maximum magnitude)\\
\textbf{Output}: Perturbation $\eta$
\begin{algorithmic}[1] 
\STATE{Random init $\eta$\,,\, $X_0=X+\eta$}
\FOR{$t$ in $T$}
\STATE{Init $G_{normgrad}=0$}
\FOR{$k$ in $K$}
\STATE{$grad \gets \nabla_{X_t} L(E_k(X_t),E_k(X_t+\eta))$}
\STATE{$grad \gets grad\,/\,\|grad\|_2$}
\STATE{$G_{normgrad} \gets\ G_{normgrad}+grad$}
\ENDFOR
\STATE{$X_t^{\prime} \gets X_t+a\,sign[G_{normgrad}]$}
\STATE{$\eta \gets clip_{\epsilon}(X_t^{\prime}-X)$}
\STATE{$X_{t+1} \gets X+\eta$}
\ENDFOR
\STATE{\textbf{return $\eta$}}
\end{algorithmic}
\end{algorithm}

\subsection{Latent Ensemble Attack}
We point out that most recent generative models can be defined as a two-stage process: (1) the latent encoding process and (2) the generation process, formulated as follows:
\begin{equation*}
y=G(E(X),c).
\end{equation*}
In the latent encoding process, the input's semantic information is encoded by the latent encoder $E$. The generator $G$ then decodes the latent representation and the target attribute to obtain the desired image. Given that $E(X)$ contains rich semantics, we demonstrate that attacking the latent encoding process can significantly mislead the generator's starting point, which makes it impossible to generate the desired image regardless of the target attributes. Based on these insights, we propose Latent Ensemble Attack (LEAT), which solely focuses on attacking the latent encoding process:
\begin{equation}\label{eq5}
\max_\eta\sum_{k}^{}L(E_k(X),E_k(X+\eta)),\,s.t.\,\|\eta\|_\infty\le\epsilon.
\end{equation}
In LEAT, the latent encoding process is independent of the target attributes, allowing for fully target attribute-agnostic attack. This means that the optimal perturbation can be obtained without relying on any target attribute throughout the entire process. Since the attack is irrelevant of the target attributes, a successful disruption in the latent space guarantees robust disruption even when unknown target attributes are given. Consequently, LEAT achieves target attribute-transferability more effectively than the Image Attack approach. Moreover, LEAT is significantly faster than Image Attack because it does not generate any output image and only focuses on attacking the latent encoding process, as illustrated in Figure \ref{fig_model}.

Specifically, LEAT extracts the latents for target models and calculates the loss between the latent for each model separately. These losses are then ensembled across the target models. Since the loss is calculated separately, we eliminate the need for any assumptions or prior knowledge about the latent representation. This enables a model-agnostic disruption that can accommodate varying shapes and semantics in latent representations. In contrast, the disruption of feature extraction module in \cite{tang2023feature} is based on an assumption that the latent representation of each model is a feature map. They aggregate the feature maps by resizing them to a fixed shape and summing them before calculating the loss. Consequently, their approach is restricted to models where the latent is in the form of a feature map, limiting the scalability to models.

\subsection{Normalized Gradient Ensemble}
To aggregate the loss across the models and compute the gradient for iterative adversarial attack, the commonly used approach is Loss Ensemble,
\begin{equation}\label{eq6}
G_{loss}=\nabla_X\sum_{k}^{}\omega_k L(M_k(X),M_k(X+\eta)),
\end{equation}
where $M_k$ can be either the latent encoder $E_k$ for LEAT or the entire model $G_k$ for Image Attack. For the Image Attack, the loss of each model is calculated as the average loss following Eq.(\ref{eq4}). However, if the weights $\omega_k$ for each model are not chosen appropriately, the Loss Ensemble approach exhibits biased attacks towards vulnerable models. To address this issue, \cite{guan2022defending} proposes Hard Model Mining (HMM), which attacks the hardest model at each iteration by updating the minimum loss among the models as follows:
\begin{equation}\label{eq7}
G_{hmm}=\nabla_X\,min\,L(M_k(X),M_k(X+\eta)).
\end{equation}
\cite{tang2023feature} points out that computing valid gradients becomes difficult when the gradients from the models are different. To obtain a better gradient direction, they propose Gradient Ensemble as follows:
\begin{equation}\label{eq8}
G_{grad}=\sum_{k}^{}\frac{1}{K}\nabla_XL(M_k(X),M_k(X+\eta)),
\end{equation}
where the gradient is computed for $K$ different models separately and then summed up.

However, we have found that both HMM \cite{guan2022defending} and Gradient Ensemble \cite{tang2023feature} remain sensitive to individual model, because their methods do not consider the relative scale differences between the models. This sensitivity hampers the scalability to target models, leading to the failure of disruption even if the loss or gradient of one particular model has a significantly different scale. Consequently, these methods primarily disrupt the vulnerable model and fail to effectively attack others. To address the issue of the scale difference misleading the gradient direction, we propose Normalized Gradient Ensemble, formulated by:
\begin{equation}\label{eq9}
G_{normgrad}=\sum_{k}^{}Norm(\nabla_X L(M_k(X),M_k(X+\eta))),
\end{equation}
where the gradient of each model is divided by its $L_2$ norm and then summed up. This ensures that the gradients are brought to the same scale, allowing all models to have an equal impact on the ensemble process. It leads to effective model-transferability, without exhibiting bias towards a particular model. We demonstrate that our simple normalization technique not only plays a key role in disrupting multiple deepfake models simultaneously, but also enhances the scalability to target models. Any model can be incorporated into the ensemble process without the concern of overwhelming its contribution. Once the gradients are aggregated, an iterative update is performed as described in Eq.(\ref{eq3}):
\begin{equation}\label{eq10}
X_{t+1}=clip(X_t+a\,sign[G_{normgrad}]).
\end{equation}
The whole process is described in Algorithm \ref{alg:algorithm}.

\section{Experiments}

In this section, we provide an overview of our implementation. We then outline the evaluation metrics we have defined. Next, we present the results of our LEAT in comparison to Image Attack. Additionally, we compare the performance of our Normalized Gradient Ensemble with previous ensemble methods. Lastly, we assess the transferability of our method in a black-box scenario.

\subsection{Implementation Details}
In our experiments, we use the CelebA-HQ \cite{karras2018progressive} dataset consisting of 30,000 high-quality facial images. From this dataset, we select 500 images as the source for protection. We employ three types of deepfake models: StyleCLIP \cite{patashnik2021styleclip} and Diffusion Autoencoders \cite{preechakul2022diffusion} for face attribute manipulation, SimSwap \cite{chen2020simswap} for face swapping, and ICface \cite{tripathy2020icface} for face reenactment. For StyleCLIP, we use five target attributes as known-targets and other five attributes as unknown-targets. Similarly, for Diffusion Autoencoders, we use two attributes respectively. For SimSwap, we select an image from CelebA-HQ as known-target face for each source, and another image as unknown-target face. In contrast to previous works \cite{guan2022defending,aneja2022tafim} that aim to protect an image from being used as a target image in SimSwap, we protect it from being used as a source image since the source affects the facial part, which is more critical to the recognition of the identity. For ICface, we use a known-target driving video from the VoxCeleb dataset \cite{nagrani17_interspeech}, as well as an unknown-target video. We extract 100 frames from each video to obtain Action Units (AUs) that guide the reenactment process.

We employ our Normalized Gradient Ensemble described in Eq.(\ref{eq9}) and Eq.(\ref{eq10}) as a default ensemble method in our experiments. In the Image Attack scenario, we generate perturbations exclusively using the known-targets, where $M_k$ represents the entire model $G_k$. Subsequently, we calculate the distance between the output images before and after disruption, utilizing the known-targets (white-box) and unknown-targets (gray-box) respectively. In our LEAT, we generate a perturbation without utilizing the target attribute information, where $M_k$ represents the latent encoder $E_k$. We then calculate the results using the same methodology as the Image Attack.

For the adversarial attack method, we employ PGD \cite{madry2018towards}, which is commonly used in previous works. We set the number of iterations $T$ to 30. During each iteration, the step size $a$ and the maximum bound $\epsilon$ are set to 0.01 and 0.05, respectively.

\begin{table}[ht]
\centering
\resizebox{0.95\columnwidth}{!}{
\begin{tabular}{c|ccc}
\hline
Models    & \makecell{Intermediate\\latent}             & \makecell{Target attribute-\\independent} & \multicolumn{1}{l}{\makecell{Low-dimensional\\semantic space}} \\ \hline
StyleCLIP & $18\times512$ latent code & \cmark          & \cmark                                \\
DiffAE    & $512\text{-}d$ vector       & \cmark          & \cmark                                \\
SimSwap   & $512\text{-}d$ vector       & \cmark          & \cmark                                \\
ICface    & Neutral Image      & \cmark          & \xmark                                \\ \hline
\end{tabular}}
\caption{Comparison of intermediate latents generated by the deepfake models.}
\label{table_latent}
\end{table}

\subsection{Evaluation Metrics}
Previous works \cite{ruiz2020disrupting,guan2022defending,tang2023feature} have commonly evaluated the effectiveness of disruption by calculating the average $L_2$ loss between output images before and after disruption. This metric, referred to as $L_2$ \textit{image}, quantifies the pixel-level difference between the images. However, as illustrated in Figure \ref{fig1}, a high $L_2$ \textit{image} value does not always indicate successful disruption. We demonstrate that capturing the semantic difference is another crucial factor for evaluating the disruption effectiveness. To assess the semantic difference, we utilize the identity loss \cite{richardson2021encoding} and LPIPS \cite{zhang2018unreasonable}. A higher identity loss implies a higher likelihood of being perceived as different individuals. LPIPS is a metric that evaluates perceptual similarity and has shown a strong correlation with human perception. Additionally, we redefine the defense success rate (DSR) to comprehensively evaluate the proportion of successful disruptions. While \cite{ruiz2020disrupting,guan2022defending} regard an attack successful when $L_2$ \textit{image} exceeds 0.05, we define success based on one of the following conditions: $L_2$ \textit{image} higher than 0.05, ID loss higher than 0.6, or LPIPS higher than 0.4. We also report Avg-DSR, which represents the average DSR of the deepfake models, and E-DSR, which measures the proportion of the protected source images that successfully disrupt all models simultaneously, following \cite{guan2022defending}.

\begin{table}[t]
\centering
\resizebox{0.85\columnwidth}{!}{
\begin{tabular}{cccccc}
\hline
\multirow{2}{*}{Models}                         & \multirow{2}{*}{Metrics $\uparrow$} & \multicolumn{2}{c}{White-box}                          & \multicolumn{2}{c}{Gray-box}      \\ \cline{3-6} 
                                                &                                     & Image           & LEAT                               & Image           & LEAT          \\ \hline
\multicolumn{1}{c|}{\multirow{4}{*}{StyleCLIP}} & \multicolumn{1}{c|}{$L_2$ \textit{image}}    & \textbf{0.4055} & \multicolumn{1}{c|}{0.0972}          & \textbf{0.2501} & 0.0816          \\
\multicolumn{1}{c|}{}                           & \multicolumn{1}{c|}{ID loss}        & 0.3516          & \multicolumn{1}{c|}{\textbf{0.4867}} & 0.3177          & \textbf{0.4699} \\
\multicolumn{1}{c|}{}                           & \multicolumn{1}{c|}{LPIPS}          & 0.5017          & \multicolumn{1}{c|}{\textbf{0.5125}} & 0.4754          & \textbf{0.5068} \\
\multicolumn{1}{c|}{}                           & \multicolumn{1}{c|}{DSR}            & \textbf{98.80\%} & \multicolumn{1}{c|}{\textbf{98.80\%}} & 94.80\%          & \textbf{98.40\%} \\ \hline
\multicolumn{1}{c|}{\multirow{4}{*}{DiffAE}}    & \multicolumn{1}{c|}{$L_2$ \textit{image}}    & \textbf{0.0578} & \multicolumn{1}{c|}{0.0312}          & \textbf{0.0502} & 0.0303          \\
\multicolumn{1}{c|}{}                           & \multicolumn{1}{c|}{ID loss}        & 0.1874          & \multicolumn{1}{c|}{\textbf{0.4208}} & 0.1724          & \textbf{0.4193} \\
\multicolumn{1}{c|}{}                           & \multicolumn{1}{c|}{LPIPS}          & \textbf{0.4491} & \multicolumn{1}{c|}{0.4390}          & \textbf{0.4509} & 0.4434          \\
\multicolumn{1}{c|}{}                           & \multicolumn{1}{c|}{DSR}            & \textbf{80.20\%} & \multicolumn{1}{c|}{57.40\%}          & \textbf{88.60\%} & 63.20\%          \\ \hline
\multicolumn{1}{c|}{\multirow{4}{*}{SimSwap}}   & \multicolumn{1}{c|}{$L_2$ \textit{image}}    & \textbf{0.0591} & \multicolumn{1}{c|}{0.0133}          & \textbf{0.0463} & 0.0133          \\
\multicolumn{1}{c|}{}                           & \multicolumn{1}{c|}{ID loss}        & 0.6066          & \multicolumn{1}{c|}{\textbf{0.9576}} & 0.5871          & \textbf{0.9538} \\
\multicolumn{1}{c|}{}                           & \multicolumn{1}{c|}{LPIPS}          & \textbf{0.1972} & \multicolumn{1}{c|}{0.1796}          & \textbf{0.1928} & 0.1812          \\
\multicolumn{1}{c|}{}                           & \multicolumn{1}{c|}{DSR}            & 84.60\%          & \multicolumn{1}{c|}{\textbf{99.80\%}} & 67.40\%          & \textbf{99.60\%} \\ \hline
\multicolumn{1}{c|}{\multirow{4}{*}{ICface}}    & \multicolumn{1}{c|}{$L_2$ \textit{image}}    & \textbf{0.1013} & \multicolumn{1}{c|}{0.1002}          & 0.0983          & \textbf{0.1031} \\
\multicolumn{1}{c|}{}                           & \multicolumn{1}{c|}{ID loss}        & 0.1925          & \multicolumn{1}{c|}{\textbf{0.1984}} & 0.1897          & \textbf{0.1992} \\
\multicolumn{1}{c|}{}                           & \multicolumn{1}{c|}{LPIPS}          & 0.2862          & \multicolumn{1}{c|}{\textbf{0.2892}} & 0.2826          & \textbf{0.2910} \\
\multicolumn{1}{c|}{}                           & \multicolumn{1}{c|}{DSR}            & \textbf{89.20\%} & \multicolumn{1}{c|}{87.40\%}          & 87.80\%          & \textbf{88.80\%} \\ \hline
\multicolumn{2}{c|}{Avg-DSR}                                                          & \textbf{88.20\%} & \multicolumn{1}{c|}{85.85\%}          & 84.65\%          & \textbf{87.50\%} \\
\multicolumn{2}{c|}{E-DSR}                                                            & \textbf{60.20\%} & \multicolumn{1}{c|}{51.00\%}          & 50.20\%          & \textbf{56.20\%} \\ \hline
\end{tabular}
}
\caption{Quantitative comparison between Image Attack and LEAT. Best results for each metric are highlighted in \textbf{bold}.}
\label{table_main}
\end{table}

\begin{figure}[t]
\centering
\includegraphics[width=0.95\columnwidth]{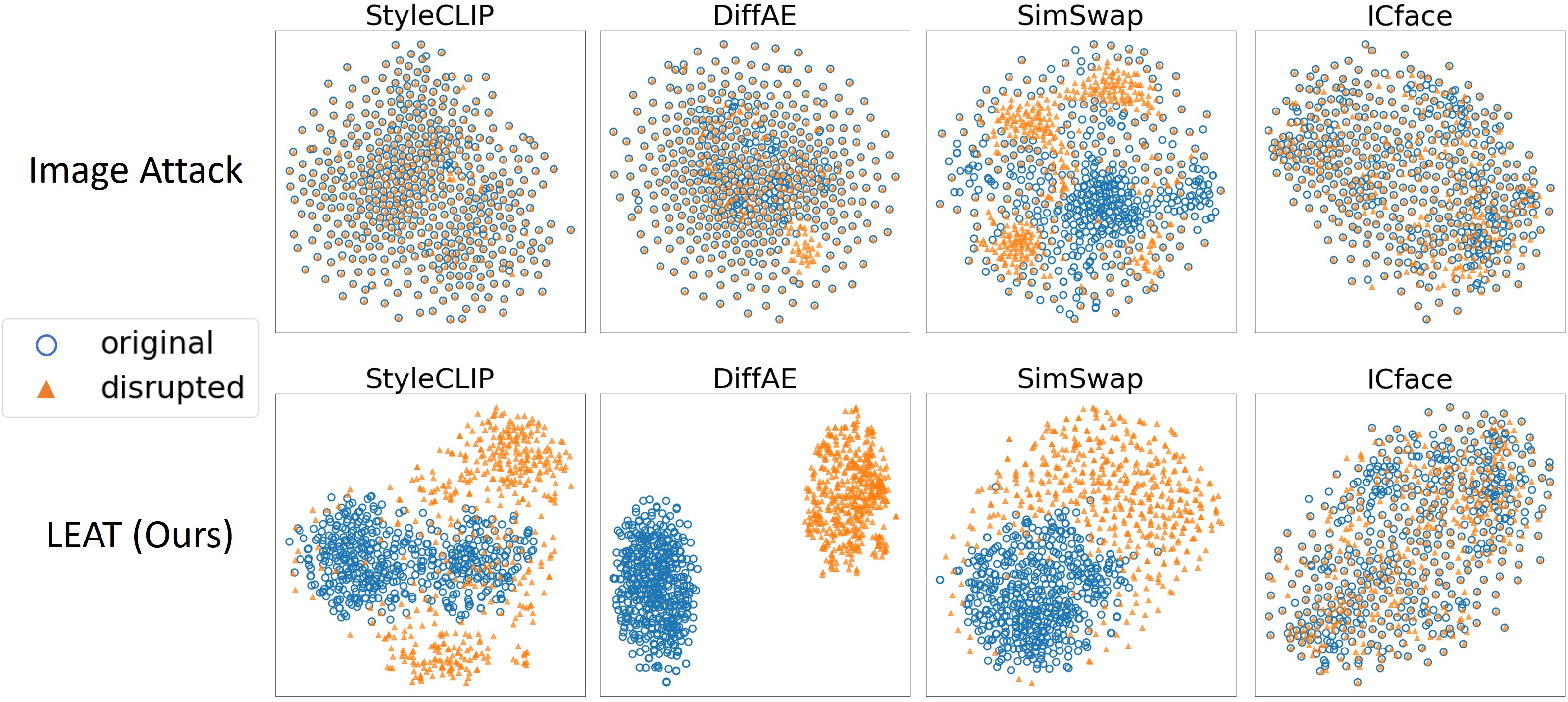}
\caption{t-SNE visualization of the original and disrupted latents under Image Attack and LEAT.}
\label{fig_tsne}
\end{figure}

\begin{figure}[t]
\centering
\includegraphics[width=0.95\columnwidth]{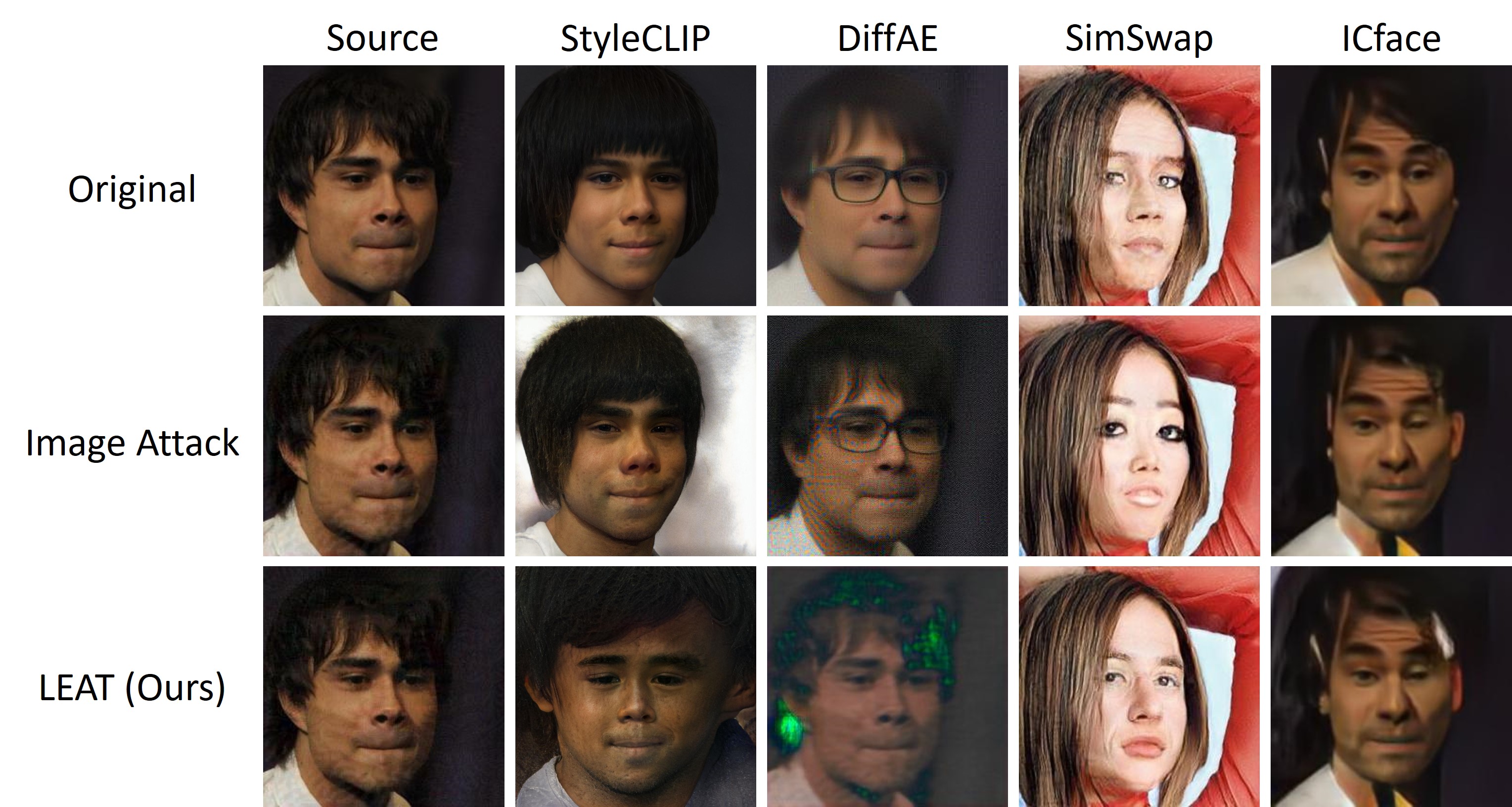}
\caption{Examples of disrupted outputs in a gray-box scenario. \textit{Top}: source and original outputs. \textit{Middle}: perturbed source and disrupted outputs by Image Attack. \textit{Bottom}: perturbed source and disrupted outputs by LEAT.}
\label{fig_main}
\end{figure}

\begin{table}[ht]
\centering
\resizebox{0.9\columnwidth}{!}{
\begin{tabular}{c|cc}
\hline
Attack Methods & Image Attack & LEAT (Ours) \\ \hline
Runtime (s)    & 254.98                 & 5.48       \\ \hline
\end{tabular}}
\caption{Runtime comparison between Image Attack and LEAT.}
\label{table_time}
\end{table}

\subsection{The Results of Latent Ensemble Attack}
To explore the impact of our LEAT, we present the distribution of the latents before and after attack using t-SNE visualization in Figure \ref{fig_tsne}. Compared to Image Attack, LEAT clearly distinguishes the latents for StyleCLIP, Diffusion Autoencoders, and SimSwap. This indicates that LEAT directs the encoded latent towards a significantly different direction. As the latent serves as a starting point for the generator, disrupted latent subsequently generates the undesired output. However, the latents of ICface are not distinguished since they are represented as neutral images, which are not embedded in a low-dimensional space. Still, they can be exploited as an attack point of LEAT due to their target attribute-independence. The properties of the latent in each model are described in Table \ref{table_latent}.

We present the quantitative results of our proposed LEAT in Table \ref{table_main}. For both LEAT and Image Attack, we employ our Normalized Gradient Ensemble as an ensemble method. In Image Attack, most of the models show a higher $L_2$ \textit{image} as it is the direct target in the PGD attack. However, in the gray-box scenario where unseen target attributes are provided, $L_2$ \textit{image} decreases significantly in Image Attack. In contrast, LEAT shows robust performance in the gray-box scenario. Additionally, LEAT consistently reports higher ID loss. In terms of LPIPS, LEAT achieves better performance in StyleCLIP and ICface. Remarkably, LEAT achieves higher Avg-DSR and E-DSR scores compared to Image Attack in the gray-box scenario, demonstrating its robust target attribute-transferability. The qualitative results of our method are shown in Figure \ref{fig_main}.

Furthermore, we compare the perturbation generation time in Table \ref{table_time} to highlight the efficiency of our LEAT. In Image Attack, the output image is generated by fully utilizing deepfake models, and the loss is averaged across all target attributes. Conversely, in LEAT, the latent is generated solely by forwarding the latent encoder, without any loss averaging from the target attributes. Consequently, perturbation generation is much faster in LEAT. With a single NVIDIA A100 GPU, Image Attack takes an average of 254.98 seconds for 500 images. In comparison, LEAT requires only 5.48 seconds under the same setting, making it approximately 46 times faster than Image Attack.

\begin{table}[t]
\centering
\resizebox{1.0\columnwidth}{!}{
\begin{tabular}{cccccccc}
\hline
\multirow{2}{*}{Models}                         & \multirow{2}{*}{Metrics $\uparrow$} & \multicolumn{3}{c}{White-box}                                            & \multicolumn{3}{c}{Gray-box}                        \\ \cline{3-8} 
                                                &                                     & Gradient        & HMM             & Ours                                 & Gradient        & HMM             & Ours            \\ \hline
\multicolumn{2}{l}{\textit{Image Attack}}                                                      &                 &                 &                                      &                 &                 &                 \\
\multicolumn{1}{c|}{\multirow{4}{*}{StyleCLIP}} & \multicolumn{1}{c|}{$L_2$ \textit{image}}    & 0.2856          & 0.0596          & \multicolumn{1}{c|}{\textbf{0.4055}} & 0.1445          & 0.0325          & \textbf{0.2501} \\
\multicolumn{1}{c|}{}                           & \multicolumn{1}{c|}{ID loss}        & 0.2267          & 0.1167          & \multicolumn{1}{c|}{\textbf{0.3516}} & 0.2067          & 0.1036          & \textbf{0.3177} \\
\multicolumn{1}{c|}{}                           & \multicolumn{1}{c|}{LPIPS}          & 0.4151          & 0.3811          & \multicolumn{1}{c|}{\textbf{0.5017}} & 0.3977          & 0.3687          & \textbf{0.4754} \\
\multicolumn{1}{c|}{}                           & \multicolumn{1}{c|}{DSR}            & 51.00\%          & 23.00\%          & \multicolumn{1}{c|}{\textbf{98.80\%}} & 38.60\%          & 13.40\%          & \textbf{94.80\%} \\ \hline
\multicolumn{1}{c|}{\multirow{4}{*}{DiffAE}}    & \multicolumn{1}{c|}{$L_2$ \textit{image}}    & 0.0282          & 0.0432          & \multicolumn{1}{c|}{\textbf{0.0578}} & 0.0237          & 0.0386          & \textbf{0.0502} \\
\multicolumn{1}{c|}{}                           & \multicolumn{1}{c|}{ID loss}        & 0.1606          & 0.1430          & \multicolumn{1}{c|}{\textbf{0.1874}} & 0.1666          & 0.1263          & \textbf{0.1724} \\
\multicolumn{1}{c|}{}                           & \multicolumn{1}{c|}{LPIPS}          & 0.3751          & 0.3982          & \multicolumn{1}{c|}{\textbf{0.4491}} & 0.3728          & 0.3990          & \textbf{0.4509} \\
\multicolumn{1}{c|}{}                           & \multicolumn{1}{c|}{DSR}            & 18.60\%          & 42.40\%          & \multicolumn{1}{c|}{\textbf{80.20\%}} & 19.60\%          & 48.80\%          & \textbf{88.60\%} \\ \hline
\multicolumn{1}{c|}{\multirow{4}{*}{SimSwap}}   & \multicolumn{1}{c|}{$L_2$ \textit{image}}    & 0.0118          & 0.0426          & \multicolumn{1}{c|}{\textbf{0.0591}} & 0.0095          & 0.0342          & \textbf{0.0463} \\
\multicolumn{1}{c|}{}                           & \multicolumn{1}{c|}{ID loss}        & 0.3114          & 0.5655          & \multicolumn{1}{c|}{\textbf{0.6066}} & 0.3011          & 0.5477          & \textbf{0.5871} \\
\multicolumn{1}{c|}{}                           & \multicolumn{1}{c|}{LPIPS}          & 0.1071          & 0.1850          & \multicolumn{1}{c|}{\textbf{0.1972}} & 0.1052          & 0.1807          & \textbf{0.1928} \\
\multicolumn{1}{c|}{}                           & \multicolumn{1}{c|}{DSR}            & 4.40\%          & 57.00\%          & \multicolumn{1}{c|}{\textbf{84.60\%}} & 2.60\%          & 42.40\%          & \textbf{67.40\%} \\ \hline
\multicolumn{1}{c|}{\multirow{4}{*}{ICface}}    & \multicolumn{1}{c|}{$L_2$ \textit{image}}    & \textbf{0.1738} & 0.0505          & \multicolumn{1}{c|}{0.1013}          & \textbf{0.1682} & 0.0498          & 0.0983          \\
\multicolumn{1}{c|}{}                           & \multicolumn{1}{c|}{ID loss}        & \textbf{0.3423} & 0.1319          & \multicolumn{1}{c|}{0.1925}          & \textbf{0.3407} & 0.1308          & 0.1897          \\
\multicolumn{1}{c|}{}                           & \multicolumn{1}{c|}{LPIPS}          & \textbf{0.3881} & 0.2282          & \multicolumn{1}{c|}{0.2862}          & \textbf{0.3841} & 0.2261          & 0.2826          \\
\multicolumn{1}{c|}{}                           & \multicolumn{1}{c|}{DSR}            & \textbf{96.80\%} & 25.80\%          & \multicolumn{1}{c|}{89.20\%}          & \textbf{95.40\%} & 26.00\%          & 87.80\%          \\ \hline
\multicolumn{2}{c|}{Avg-DSR}                                                          & 42.70\%          & 37.05\%          & \multicolumn{1}{c|}{\textbf{88.20\%}} & 39.05\%          & 32.65\%          & \textbf{84.65\%} \\
\multicolumn{2}{c|}{E-DSR}                                                            & 1.00\%          & 6.20\%          & \multicolumn{1}{c|}{\textbf{60.20\%}} & 0.60\%          & 1.20\%          & \textbf{50.20\%} \\ \hline
\multicolumn{1}{l}{\textit{LEAT}}               &                                     &                 &                 &                                      &                 &                 &                 \\
\multicolumn{1}{c|}{\multirow{4}{*}{StyleCLIP}} & \multicolumn{1}{c|}{$L_2$ \textit{image}}    & 0.0118          & 0.0074          & \multicolumn{1}{c|}{\textbf{0.0972}} & 0.0099          & 0.0067          & \textbf{0.0816} \\
\multicolumn{1}{c|}{}                           & \multicolumn{1}{c|}{ID loss}        & 0.0462          & 0.0322          & \multicolumn{1}{c|}{\textbf{0.4867}} & 0.0428          & 0.0306          & \textbf{0.4699} \\
\multicolumn{1}{c|}{}                           & \multicolumn{1}{c|}{LPIPS}          & 0.2693          & 0.2761          & \multicolumn{1}{c|}{\textbf{0.5125}} & 0.2725          & 0.2808          & \textbf{0.5068} \\
\multicolumn{1}{c|}{}                           & \multicolumn{1}{c|}{DSR}            & 0.60\%          & 0.40\%          & \multicolumn{1}{c|}{\textbf{98.80\%}} & 0.60\%          & 0.40\%          & \textbf{98.40\%} \\ \hline
\multicolumn{1}{c|}{\multirow{4}{*}{DiffAE}}    & \multicolumn{1}{c|}{$L_2$ \textit{image}}    & \textbf{0.0569} & 0.0196          & \multicolumn{1}{c|}{0.0312}          & \textbf{0.0535} & 0.0183          & 0.0303          \\
\multicolumn{1}{c|}{}                           & \multicolumn{1}{c|}{ID loss}        & \textbf{0.5253} & 0.0863          & \multicolumn{1}{c|}{0.4208}          & \textbf{0.5045} & 0.0815          & 0.4193          \\
\multicolumn{1}{c|}{}                           & \multicolumn{1}{c|}{LPIPS}          & \textbf{0.5608} & 0.2893          & \multicolumn{1}{c|}{0.4390}          & \textbf{0.5647} & 0.2927          & 0.4434          \\
\multicolumn{1}{c|}{}                           & \multicolumn{1}{c|}{DSR}            & \textbf{98.20\%} & 0.40\%          & \multicolumn{1}{c|}{57.40\%}          & \textbf{98.20\%} & 0.20\%          & 63.20\%          \\ \hline
\multicolumn{1}{c|}{\multirow{4}{*}{SimSwap}}   & \multicolumn{1}{c|}{$L_2$ \textit{image}}    & 0.0011          & \textbf{0.0148} & \multicolumn{1}{c|}{0.0133}          & 0.0011          & \textbf{0.0148} & 0.0133          \\
\multicolumn{1}{c|}{}                           & \multicolumn{1}{c|}{ID loss}        & 0.0660          & \textbf{1.1034} & \multicolumn{1}{c|}{0.9576}          & 0.0661          & \textbf{1.1011} & 0.9538          \\
\multicolumn{1}{c|}{}                           & \multicolumn{1}{c|}{LPIPS}          & 0.0331          & \textbf{0.1914} & \multicolumn{1}{c|}{0.1796}          & 0.0333          & \textbf{0.1929} & 0.1812          \\
\multicolumn{1}{c|}{}                           & \multicolumn{1}{c|}{DSR}            & 0.00\%          & \textbf{100.00\%} & \multicolumn{1}{c|}{99.80\%}          & 0.00\%          & \textbf{99.80\%} & 99.60\%          \\ \hline
\multicolumn{1}{c|}{\multirow{4}{*}{ICface}}    & \multicolumn{1}{c|}{$L_2$ \textit{image}}    & 0.0019          & 0.0102          & \multicolumn{1}{c|}{\textbf{0.1002}} & 0.0019          & 0.0105          & \textbf{0.1031} \\
\multicolumn{1}{c|}{}                           & \multicolumn{1}{c|}{ID loss}        & 0.0209          & 0.0362          & \multicolumn{1}{c|}{\textbf{0.1984}} & 0.0211          & 0.0370          & \textbf{0.1992} \\
\multicolumn{1}{c|}{}                           & \multicolumn{1}{c|}{LPIPS}          & 0.0401          & 0.0903          & \multicolumn{1}{c|}{\textbf{0.2892}} & 0.0404          & 0.0909          & \textbf{0.2910} \\
\multicolumn{1}{c|}{}                           & \multicolumn{1}{c|}{DSR}            & 0.00\%          & 0.00\%          & \multicolumn{1}{c|}{\textbf{87.40\%}} & 0.00\%          & 0.00\%          & \textbf{88.80\%} \\ \hline
\multicolumn{2}{c|}{Avg-DSR}                                                          & 24.70\%          & 25.20\%          & \multicolumn{1}{c|}{\textbf{85.85\%}} & 24.70\%          & 25.10\%          & \textbf{87.50\%} \\
\multicolumn{2}{c|}{E-DSR}                                                            & 0.00\%          & 0.00\%          & \multicolumn{1}{c|}{\textbf{51.00\%}} & 0.00\%          & 0.00\%          & \textbf{56.20\%} \\ \hline
\end{tabular}
}
\caption{Comparison of ensemble methods applied to Image Attack and LEAT.}
\label{table_main2}
\end{table}

\subsection{Comparison with Previous Ensemble Methods}
\begin{figure}[t]
\centering
\includegraphics[width=0.95\columnwidth]{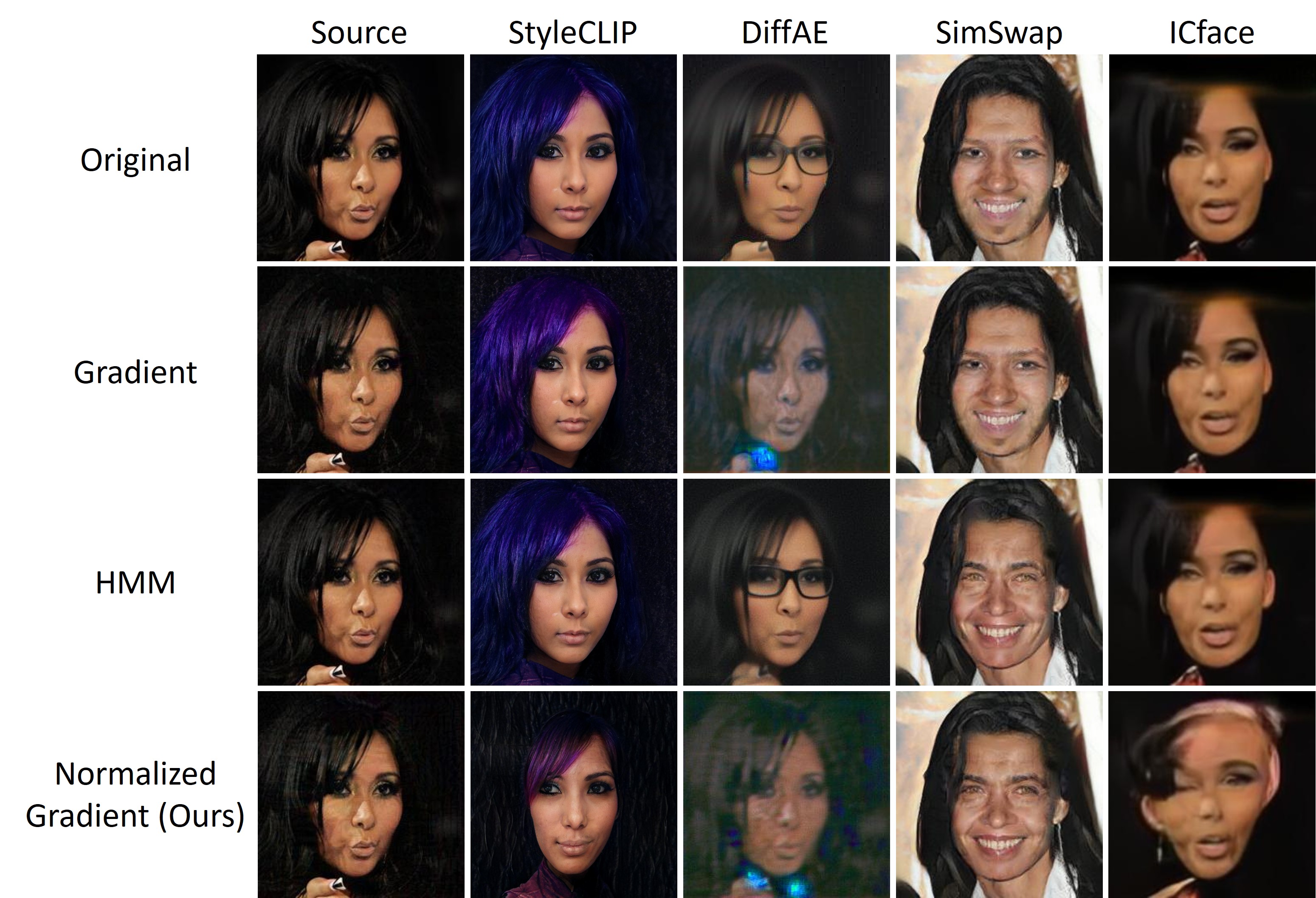}
\caption{Examples of disrupted outputs by applying different ensemble methods to LEAT in a gray-box scenario.}
\label{fig_main2}
\end{figure}

To evaluate the effectiveness of our proposed Normalized Gradient Ensemble, we compare our method with Hard Model Mining \cite{guan2022defending} and Gradient Ensemble \cite{tang2023feature} proposed in previous studies. For their methods, we follow the process outlined in Eq.(\ref{eq7}) and Eq.(\ref{eq8}), respectively. For a fair comparison, we apply each method to both Image Attack and LEAT. The quantitative and qualitative results are reported in Table \ref{table_main2} and Figure \ref{fig_main2}. In both Image Attack and LEAT, Normalized Gradient Ensemble demonstrates strong model-transferability, reporting significantly higher Avg-DSR and E-DSR scores. In contrast, both Gradient Ensemble and HMM show lower Avg-DSR and nearly zero E-DSR scores, indicating a lack of successful disruption across all models simultaneously. Specifically, Gradient Ensemble exhibits biased results towards ICface in Image Attack, leading to poor performance in the other models. In LEAT, it exclusively attacks Diffusion Autoencoders, resulting in clear disruption for Diffusion Autoencoders while leaving the others unchanged, as depicted in Figure \ref{fig_main2} (Gradient). Similarly, HMM demonstrates a strong bias to SimSwap in LEAT.

\begin{table}[ht]
\centering
\resizebox{0.9\columnwidth}{!}{
\begin{tabular}{c|ccc}
\hline
Metrics $\uparrow$ & Random & Image Attack & LEAT (Ours)             \\ \hline
$L_2$ \textit{image}        & 0.0213 & 0.0365       & \textbf{0.0409} \\
ID loss            & 0.1565 & 0.2887       & \textbf{0.3482} \\
LPIPS              & 0.2214 & 0.3458       & \textbf{0.3795} \\ \hline
\end{tabular}
}
\caption{Comparison of disruption methods in a black-box scenario against StarGAN.}
\label{table_black}
\end{table}

\subsection{Experiments in a Black-Box Scenario}
\begin{figure}[ht]
\centering
\includegraphics[width=0.9\columnwidth]{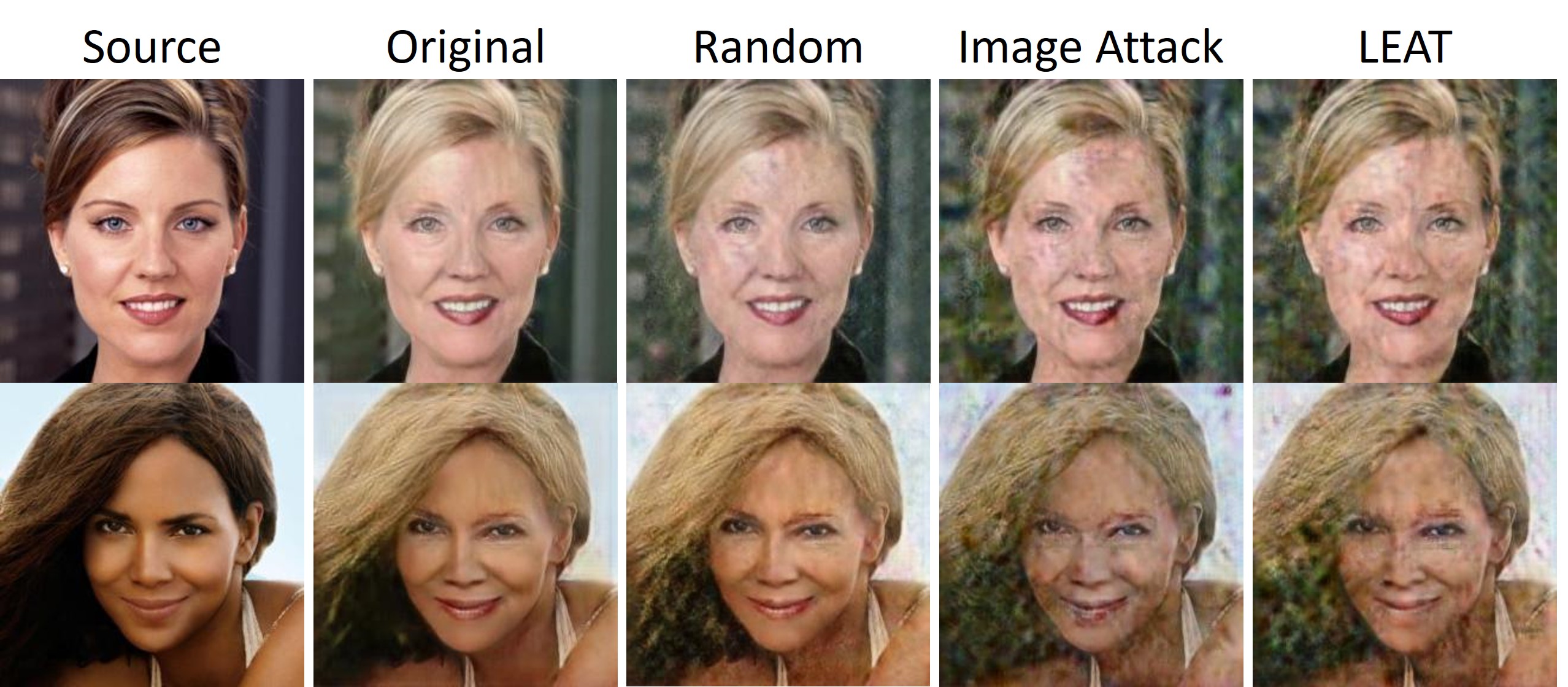}
\caption{Examples of disrupted outputs using different attack methods in a black-box scenario.}
\label{fig_black}
\end{figure}

To demonstrate the robust model-transferability of our disruption method under challenging conditions, we conduct experiments in a black-box scenario where the deepfake model is unknown. We select StarGAN \cite{choi2018stargan} as the unknown target model and use random Gaussian noise, Image Attack, and LEAT, all applied at the same scale of the perturbation to disrupt StarGAN. For each attack method, we generate a perturbation on four known-models and directly apply the perturbation to disrupt StarGAN. The quantitative results averaged over 500 images are in Table \ref{table_black}. Overall, LEAT outperforms random perturbation and Image Attack, demonstrating its strong model-transferability in the black-box scenario. The qualitative results are shown in Figure \ref{fig_black}.

\section{Conclusion}
In this paper, we propose a fully target attribute-agnostic approach called Latent Ensemble Attack, which aims to disrupt deepfake models by generating effective perturbations. Unlike previous methods that focus on maximizing the difference between generated images, our approach targets the latent encoding process to ensure disruption during subsequent generation process, leading to robust target attribute-transferability. Additionally, we introduce the Normalized Gradient Ensemble, a technique to aggregate losses from the multiple models including both GAN-based and Diffusion-based models. By uniformly scaling the gradients, we prevent the ensemble attack from exhibiting bias towards any specific model and achieve high model-transferability. Our proposed method demonstrates strong robustness in the gray-box scenario, where the target attributes are unknown. Moreover, it can be effectively applied even in a black-box scenario, where the specific deepfake model is unidentified. These results highlight the versatility and effectiveness of our approach in real-world scenarios.

\bibliography{aaai23}

\end{document}